\documentclass{article}

\usepackage{arxiv}

\usepackage[utf8]{inputenc} % allow utf-8 input
\usepackage[T1]{fontenc}    % use 8-bit T1 fonts
\usepackage{hyperref}       % hyperlinks
\usepackage{url}            % simple URL typesetting
\usepackage{booktabs}       % professional-quality tables
\usepackage{amsfonts}       % blackboard math symbols
\usepackage{nicefrac}       % compact symbols for 1/2, etc.
\usepackage{microtype}      % microtypography
\usepackage{lipsum}		% Can be removed after putting your text content
\usepackage{graphicx}
\usepackage[authoryear,round]{natbib}
\usepackage{doi}

\title{Illuminating Spaces\\ Deep Reinforcement Learning and Laser-Wall Partitioning for Architectural Layout Generation}

\author{ \href{https://orcid.org/0000-0002-2946-9668}{\includegraphics[scale=0.06]{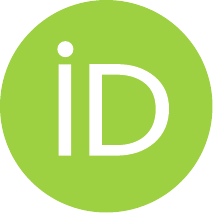}\hspace{1mm}Reza Kakooee} \\
	Institute of Technology in Architecture,\\
        Department of Architecture,\\ 
        ETH Zurich, Switzerland\\
	\texttt{kakooeer@ethz.ch} \\
	\And
	\href{https://orcid.org/0000-0002-5153-2985}{\includegraphics[scale=0.06]{orcid.pdf}\hspace{1mm}Benjamin Dillenburger} \\
	Institute of Technology in Architecture,\\
        Department of Architecture,\\ 
        ETH Zurich, Switzerland\\
}

\renewcommand{\shorttitle}{Illuminating Spaces}

\hypersetup{
pdftitle={Illuminating Spaces\\ Deep Reinforcement Learning and Laser-Wall Partitioning for Architectural Layout Generation},
pdfauthor={Reza Kakooee, Benjamin Dillenburger},
}

\begin{document}
\maketitle

\begin{abstract}
Space layout design (SLD), occurring in the early stages of the design process, nonetheless influences both the functionality and aesthetics of the ultimate architectural outcome. The complexity of SLD necessitates innovative approaches to efficiently explore vast solution spaces. While image-based generative AI has emerged as a potential solution, they often rely on pixel-based space composition methods that lack intuitive representation of architectural processes. This paper leverages deep Reinforcement Learning (RL), as it offers a procedural approach that intuitively mimics the process of human designers. Effectively using RL for SLD requires an explorative space composing method to generate desirable design solutions. We introduce "laser-wall", a novel space partitioning method that conceptualizes walls as emitters of imaginary light beams to partition spaces. This approach bridges vector-based and pixel-based partitioning methods, offering both flexibility and exploratory power in generating diverse layouts. We present two planning strategies: one-shot planning, which generates entire layouts in a single pass, and dynamic planning, which allows for adaptive refinement by continuously transforming laser-walls. Additionally, we introduce on-light and off-light wall transformations for smooth and fast layout refinement, as well as identity-less and identity-full walls for versatile room assignment. We developed SpaceLayoutGym, an open-source OpenAI Gym compatible simulator for generating and evaluating space layouts. The RL agent processes the input design scenarios and generates solutions following a reward function that balances geometrical and topological requirements. Our results demonstrate that the RL-based laser-wall approach can generate diverse and functional space layouts that satisfy both geometric constraints and topological requirements and is architecturally intuitive. 
\end{abstract}

% keywords can be removed
\keywords{Deep Reinforcement Learning, Space Layout Design, Computational Design}

\section{Introduction}
SLD represents a complex optimization problem that permeates various domains of human endeavor. At its core, SLD involves the strategic arrangement of elements within a defined spatial boundary to achieve specific objectives from maximizing usability to ensuring aesthetic harmony. This process necessitates a delicate balance between functional requirements including geometrical constraints, and topological relationships, often leading to the exploration of vast solution spaces in search of optimal configurations \citep{michalek2002architectural}.

SLD challenges span diverse applications, forming the foundation of efficient building design in architecture and influencing residential to complex commercial structures \citep{Lobos}. Urban planners apply SLD principles to shape city layouts, impacting urban livability and sustainability \citep{mourshed2009automated}, while interior designers use SLD to optimize furniture arrangement for functionality and user experience \citep{saruwono2012living}. In manufacturing, SLD enhances factory layouts, boosting productivity and safety \citep{drira2007facility}, and in technology, it is crucial for component arrangement in integrated circuits, balancing performance and heat dissipation \citep{kahng2011vlsi}. Game designers also employ SLD to craft engaging virtual environments \citep{liu2021virtual}. While SLD has diverse applications, this research focuses on architecture due to its impact on daily life, well-being, and societal structures, nonetheless, the proposed methodology can be extended to other domains.

Despite its widespread application, achieving optimal SLD solutions remains a challenge due to the need for advanced optimization techniques \citep{medjdoub2000separating}. The complexity of SLD stems from its inherently combinatorial nature and the need to satisfy multiple, often conflicting objectives simultaneously. Automated layout design has been addressed through various computational methods. Traditional methods such as shape grammars \citep{stiny1978palladian}, use rule-based systems to generate designs, while graph-based methods represent spatial relationships for topological analysis \citep{levin1964use, grasl2013topologies, shekhawat2020gplan, march2020geometry}. Evolutionary algorithms have laid the groundwork for systematic exploration of design possibilities while adhering to architectural principles and constraints \citep{michalek2002architectural, jo1998space, thakur2010architectural, rodrigues2013approach}. However, these approaches often fall short in capturing the intricate nuances of architectural design.

Machine learning methods have also contributed to automating and optimizing SLD \citep{merrell2010computer, dillenburger2016raumindex, chaillou2019archigan, mostafavi2024interactive}. House-GAN++ \citep{nauata2021house} has demonstrated the ability to generate house layouts in pixel-level. HouseDiffusion \citep{shabani2023housediffusion} employs denoising processes to enable precise control over vector-based floor plans, inspired by advancements in diffusion techniques \citep{NEURIPS20204c5bcfec}. Researchers have also explored the potential of RL in various aspects of architecture. \citep{wang2021artificial} developed an approach that combines RL with multi-agent systems to evolve architectural forms, addressing the limitations of conventional parametric design methods. \citep{kim2020spatial} used RL algorithms to optimize building layouts, while \citep{ren2020spatial} applied RL to adjust 3D room configurations. \citep{tigasspatial} created a self-play RL framework to optimize spatial structure assembly. \citep{ruiz2013design} and \citep{mandow2020architectural} integrated RL with shape grammars for housing layouts, while \citep{shi2020addressing} applied RL to generate simple floor plans. RL has also been used to optimize factory layouts \citep{klar2021implementation} and furniture arrangements \citep{di2021multi}. Recent work by \citep{veloso2023spatial} highlights the potential of multi-agent RL in complex spatial synthesis across various scales.

However, despite these advancements, many AI methods still rely on pixel-based methods that lack intuitive representation, are limited in addressing design requirements, or often fail to capture the nuanced interplay between form and context in architectural design. This limitation emphasizes the need for more procedurally driven methods that are flexible, efficient, and intuitive. This paper introduces a novel space composition method called "laser-wall” partitioning, which conceptualizes walls as emitters of imaginary light to divide spaces. This partitioning method allows us to leverage RL for architectural SLD, providing a procedural method that mirrors human design processes while maintaining flexibility in exploring spatial configurations.
\vspace{-5pt}
\section{Methodology} \vspace{-8pt}
Architectural SLD involves a synergy of space composition methods and computational approaches to optimize the composition. In this paper, we focus on partitioning as the space composition method and RL as a computational approach to optimize layout composition given a design scenario.

In the following section, we introduce a novel space partitioning method called "laser-wall". This innovative method enhances the effectiveness of subsequent computational approaches, such as RL, in exploring the solution space to identify optimal layouts. Moreover, the laser-wall method transforms the layout design problem into a game-like environment, making it particularly amenable to RL algorithms, leveraging their proven efficacy in finding optimal policies. We then use RL to find optimal solutions for the optimal space layout using our proposed composition methods.

\subsection{Laser-wall partitioning}
Space partitioning can be broadly categorized into two main methods: vector-based and pixel-based. Rectangular dissection, a common vector method, divides space using straight lines but often results in long, inflexible walls, limiting irregular shapes. Pixel-based partitioning, in contrast, allows for more flexible and irregular designs by dividing space at a granular level but is slower and computationally intensive. These contrasting approaches highlight the need for a more balanced method, leading to the development of our laser-wall partitioning technique, which aims to combine the strengths of both.

\subsubsection{Basics of laser-wall method}
A laser-wall consists of two components: a hard part (the base wall) and a soft part (light beams). The base wall is formed by two connected segments, either aligned or perpendicular, establishing the physical structure. From each end of these segments, light beams are emitted. These beams travel through the space, defining partitions and interacting with other elements in the layout. When a base wall is placed within the plan, its light beams automatically emit and propagate until they encounter obstacles; either the outline of the plan, other base walls, or other light beams. A key concept governing beam interactions is the infiltration rate, which introduces flexibility to the partitioning process. We explore two implementations: fixed and decreasing infiltration rate. In the former, all walls have the same infiltration rate; beams stop upon encountering other beams. In the latter, the infiltration rate decreases with distance from the beam's source, allowing beams closer to their origin to cut through beams from other walls if they have a lower infiltration rate at the intersection point. 

\subsection{Planning Approaches}
\begin{figure}
	\centering
    \includegraphics[width=0.8\linewidth]{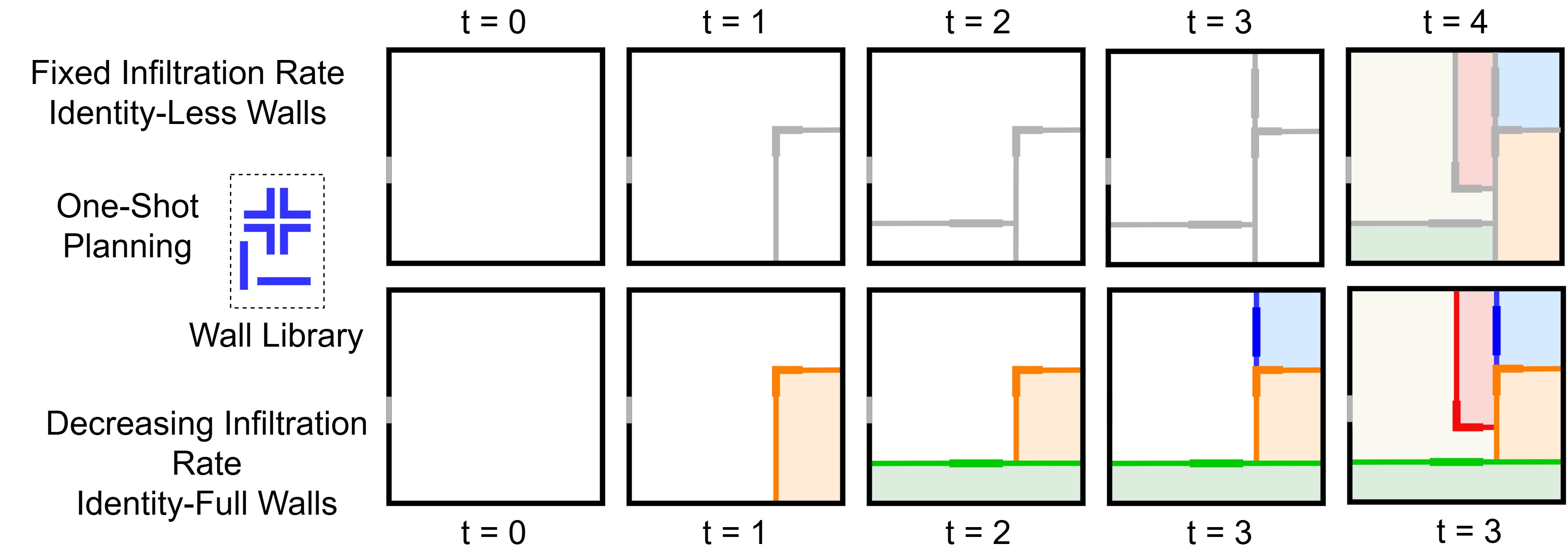}
	\caption{Comparison of one-shot planning approaches using fixed infiltration rate with identity-less walls (top row) and decreasing infiltration rate with identity-full walls (bottom row)}
	\label{fig:fig1}
\end{figure}

Building on the laser-wall concept, we introduce two planning approaches: One-Shot Planning, and Dynamic Planning.
In one-shot planning, the entire layout is determined in a single iteration. Starting with an empty plan, base laser-walls are selected from a predefined wall library and placed at chosen coordinates. Upon placement, their light beams are emitted to partition the space. This process continues until the required number of rooms is achieved. Once placed, walls remain fixed, completing the layout in one iteration. Two mechanisms are proposed for assigning rooms to subregions. In the identity-less mechanism, walls are not linked to specific rooms during partitioning. After partitioning, the subregion connected to the entrance is designated as the living room, and other rooms are assigned by matching each subregion to the room with the closest desired area defined in design scenario. In contrast, the identity-full mechanism assigns each wall a room index before placement. The smaller subregion created by the wall is assigned to its associated room, with constraints ensuring walls are not placed in already occupied regions or the entrance-connected area reserved for the living room. This process is exemplified in Figure \ref{fig:fig1}

Dynamic planning introduces adaptability by allowing walls to transform within the plan, enabling exploration of a wider range of spatial configurations. Walls can move and rotate: through incremental movements, walls shift in 8 cardinal directions (north, northeast, east, etc.) to refine their positions, while 90-degree rotations allow entire base walls or individual segments to turn clockwise or counterclockwise, altering the shape and proportions of rooms. This discrete approach aligns with a cell-based representation of the plan, where space is divided into a grid of units.

\begin{figure}
	\centering
	\includegraphics[width=0.8\linewidth]{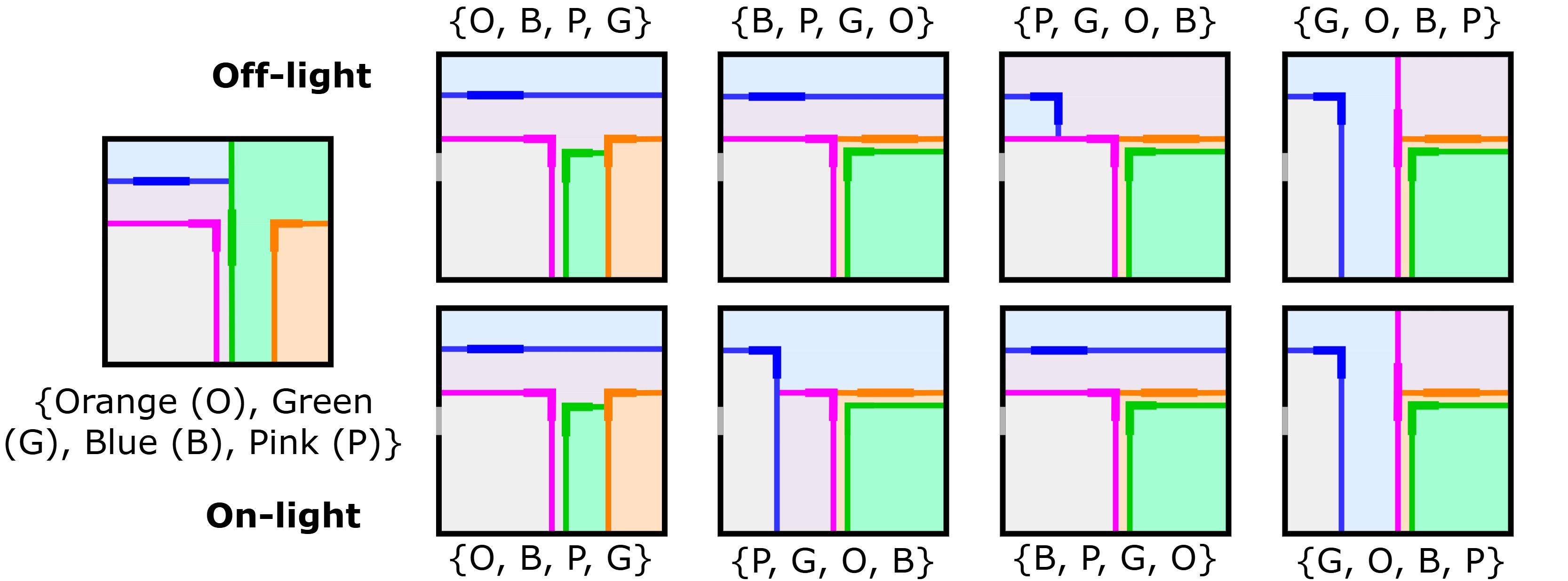}
	\caption{Illustration of off-light and on-light transformations in laser-wall partitioning}
	\label{fig:fig2}
\end{figure}

Wall movement requires determining how lights will behave during transformations. Two types of wall transformations are introduced: off-light and on-light transformations. In the former, the light of the moving wall is deactivated before its movement. After the wall is repositioned, all walls' lights are activated in a specific order, with the moved wall's light being activated last. In the latter, the light of the moving wall stays active throughout its movement. This process is practically similar to initially turning off all lights and then incrementally activating them in order, with the moved wall’s light being turned on last. After each transformation and following partitioning, first, connected subregions to each base-wall are identified, next, Intersection over Union (IoU) score between new subregions and their previous counterparts are calculated, and then rooms are assigned based on the following steps: a) the region connected to the entrance is assigned as the living room, b) subregions with an IoU of 1 (perfect overlap) are assigned to maintain continuity, c) remaining regions are assigned by iterating through walls in a specific order (with the moved wall considered last), assigning regions with the highest IoU to corresponding walls. This process is exemplified in Figure \ref{fig:fig2}.

According to the laser-wall partitioning method, we developed SpaceLayoutGym (SLG), an OpenAI Gym compatible simulator. It supports both one-shot and dynamic planning approaches, allowing exploration of various SLD strategies. SLG provides a standardized environment for creating design scenarios, evaluating layouts, and optimizing solutions based on multiple criteria. 

In this paper, we focus exclusively on dynamic planning, building upon our previous work on one-shot planning \citep{kakooee2024reimagining}. Our investigation examines dynamic planning using on-light and off-light transformations to compare the resulting layouts. The goal is not to identify a "best" method, but to understand how these variations impact agent performance and layout characteristics, offering insights into the flexibility of our RL-based approach to SLD. We also analyze how different wall types—angled, straight, or both—affect the RL agent's performance across different scenarios. 

\subsection{RL for SLD}
RL provides a framework for an agent to learn optimal decision-making strategies through interaction with an environment, making it well-suited to the iterative process of layout refinement in dynamic planning. To apply RL to our SDL with the laser-wall partitioning method, we frame the SLD problem as a Markov Decision Process (MDP). 

An MDP provides a framework for modeling decision-making in scenarios where outcomes are partly under the control of a decision-maker. The MDP is defined by three components, state space, action space, and reward function.

\subsection{State Space}
The state space in RL represents all possible configurations or situations the agent can encounter in the environment. In SLG, the state at any given time is represented by an RGB image of the current layout. This image encapsulates both geometric and topological information, including room configurations, wall positions, and the entrance location. The use of an image state allows the agent to perceive spatial relationships intuitively, mirroring human visual processing of layouts.

\subsection{Action Space}
The action space in RL defines the set of all possible actions an agent can take to interact with or modify the environment. In dynamic planning, actions consist of two decisions: a) selecting an existing wall to transform, b) choosing a specific transformation (8 directional movements and 6 rotation options).

\subsection{Reward Function}
The reward function in RL quantifies the desirability of actions and resulting states, guiding the agent towards optimal behavior. In dynamic planning, each action results in a complete new configuration, allowing for assessment of both geometric and topological properties at every step. The reward consists of two components: instant reward, and terminal reward. The reward is calculated from two elements: an instant reward and a terminal reward.\\\\

\textbf{Instant rewards}
\begin{itemize}
    \item Negative rewards for violating hard constraints (e.g., wall collisions, entrance blockage).
    \item Negative rewards for intermediate layouts that deviate significantly from desired properties.
\end{itemize}

\textbf{Terminal rewards}
\begin{itemize}
    \item Negative rewards if the layout is significantly different from the desired properties.
    \item Non-negative reward for solutions that fall within a threshold of closeness to desired properties, scaled linearly or nonlinearly based on the closeness.
    \item Bonus reward for perfect adjacency matching.
\end{itemize}

\subsection{Learning Algorithm} 
We employ Proximal Policy Optimization (PPO) \citep{schulman2017proximal} algorithm, to learn an optimal policy for layout generation. This choice is motivated by PPO’s ability to handle high-dimensional state spaces and its stability in learning complex strategies. The neural network architecture underlying our RL agent is designed to process the image-based state efficiently. It combines convolutional layers to extract spatial features, following by fully connected layers to map these features to action probabilities and state value estimates.

\subsection{Training Process} 
The agent learns through episodic interaction with the SLG environment. Each episode in dynamic planning begins with an initial random wall configuration and terminates when either a satisfactory layout is achieved or a maximum number of transformations is reached. The agent gradually improves its policy by maximizing cumulative rewards across episodes.
This RL formulation allows our system to learn design strategies that balance local optimizations (individual room properties) with global constraints (overall layout coherence and adjacency requirements). 

\section{Experiments}
This section presents two experiments to evaluate the effectiveness of laser-wall partitioning in SLG and the performance of RL to design space layouts. We define six distinct design scenarios, varying in complexity from 4 to 9 rooms. Table 1 outlines the properties of the desired layout for each scenario. For all scenarios, the desired room aspect ratio ranges from $1$ to $6$. The desired adjacency requirements are as follows: living room is connected to all the other rooms and to the entrance; each room access to at least one non-blocked façade.

\begin{table}[h]
    \centering
    \caption{Design Scenarios Specifications}
    \begin{tabular}{cccc}
        \toprule
        Scenario Number & Number of Rooms & Desired Areas (in cell counts) & Entrance Location \\
        \midrule
        1 & 4 & (431, 322, 250, 206) & West \\
        2 & 5 & (301, 220, 220, 214, 210) & East \\
        3 & 6 & (279, 220, 186, 160, 158, 198) & West \\
        4 & 7 & (313, 150, 144, 144, 130, 138, 134) & West \\
        5 & 8 & (299, 216, 210, 192, 183, 170, 160, 150) & South \\
        6 & 9 & (435, 228, 190, 178, 150, 146, 140, 140, 134) & East \\
        \bottomrule
    \end{tabular}
    \label{tab:scenario_data}
\end{table}
 
We conducted two experiments a) assessing the impact of wall type flexibility by comparing layouts generated with straight versus angled walls; and b) comparing on-light and off-light transformations to evaluate their impact on layout generation.

\subsection{Results}
\begin{figure}
	\centering
	\includegraphics[width=\linewidth]{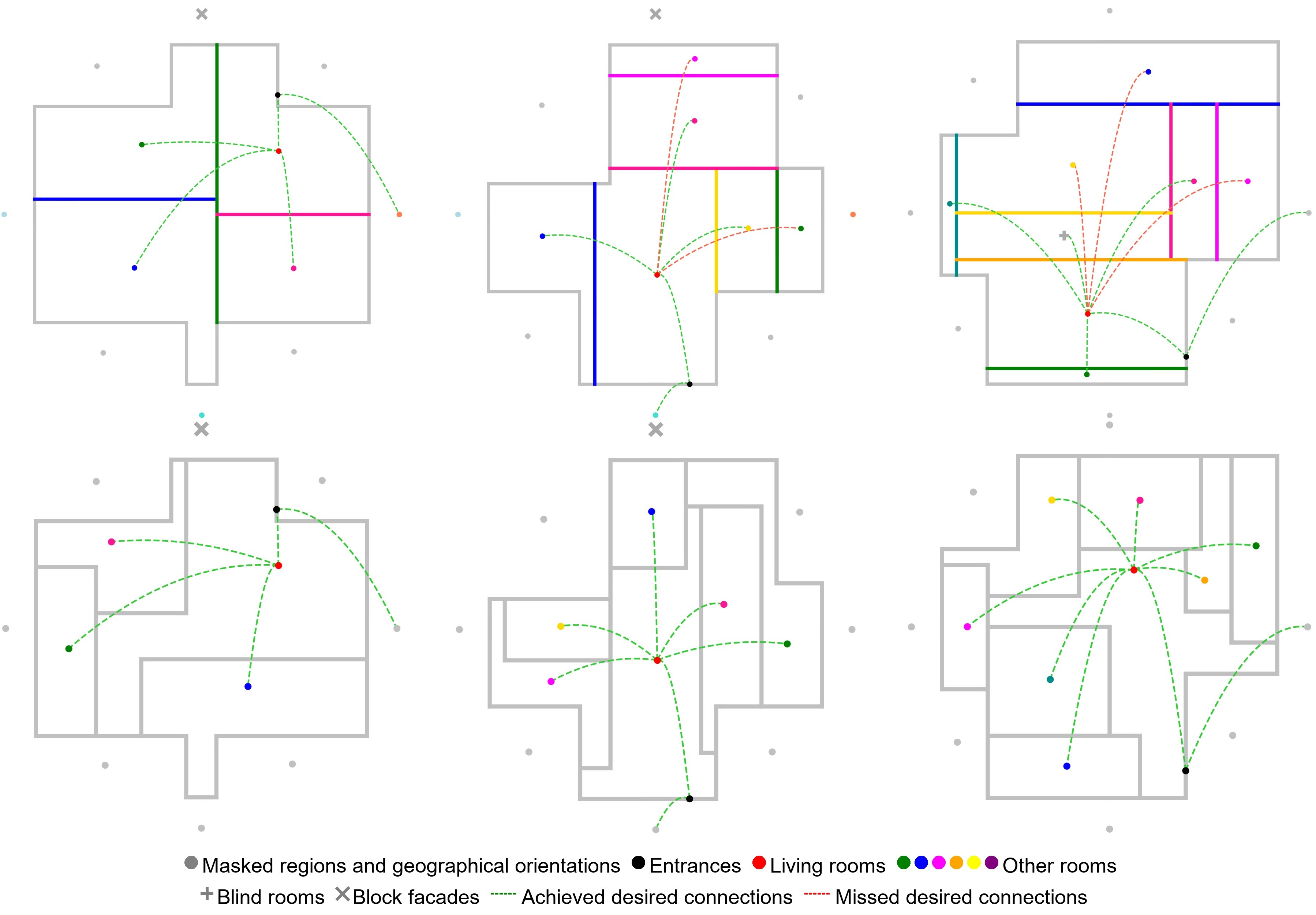}
	\caption{Comparison of layouts generated using straight walls only (top row) and angled walls only (bottom row).}
	\label{fig:fig3}
\end{figure}

Angled versus straight walls: Figure \ref{fig:fig3} shows the results for training PPO agents on a subset of design scenarios from Table 1, comparing layouts generated using only straight walls versus those using only angled walls. The agent's ability to find a design solution indicates that the generated layouts closely match the geometric properties defined in the table, including the aspect ratio requirements. This is because geometric properties are integral constraints of the optimization problem, and without satisfying these constraints, a solution cannot be achieved. Therefore, the immediate conclusion is that the PPO agents were able to find solutions that closely match the desired geometric properties. However, observing the dashed lines on the layout reveals numerous missed connections with straight walls, whereas layouts generated with angled walls show marked improvement, successfully establishing all required connections. This demonstrates that using angled walls leads to more effective space partitioning, emphasizing that combining both straight and angled walls is promising, and highlighting the strength of our laser-wall method compared to conventional rectangular dissection.

\subsubsection{On-light and off-light transformations}
\begin{figure}
	\centering
	\includegraphics[width=0.6\linewidth]{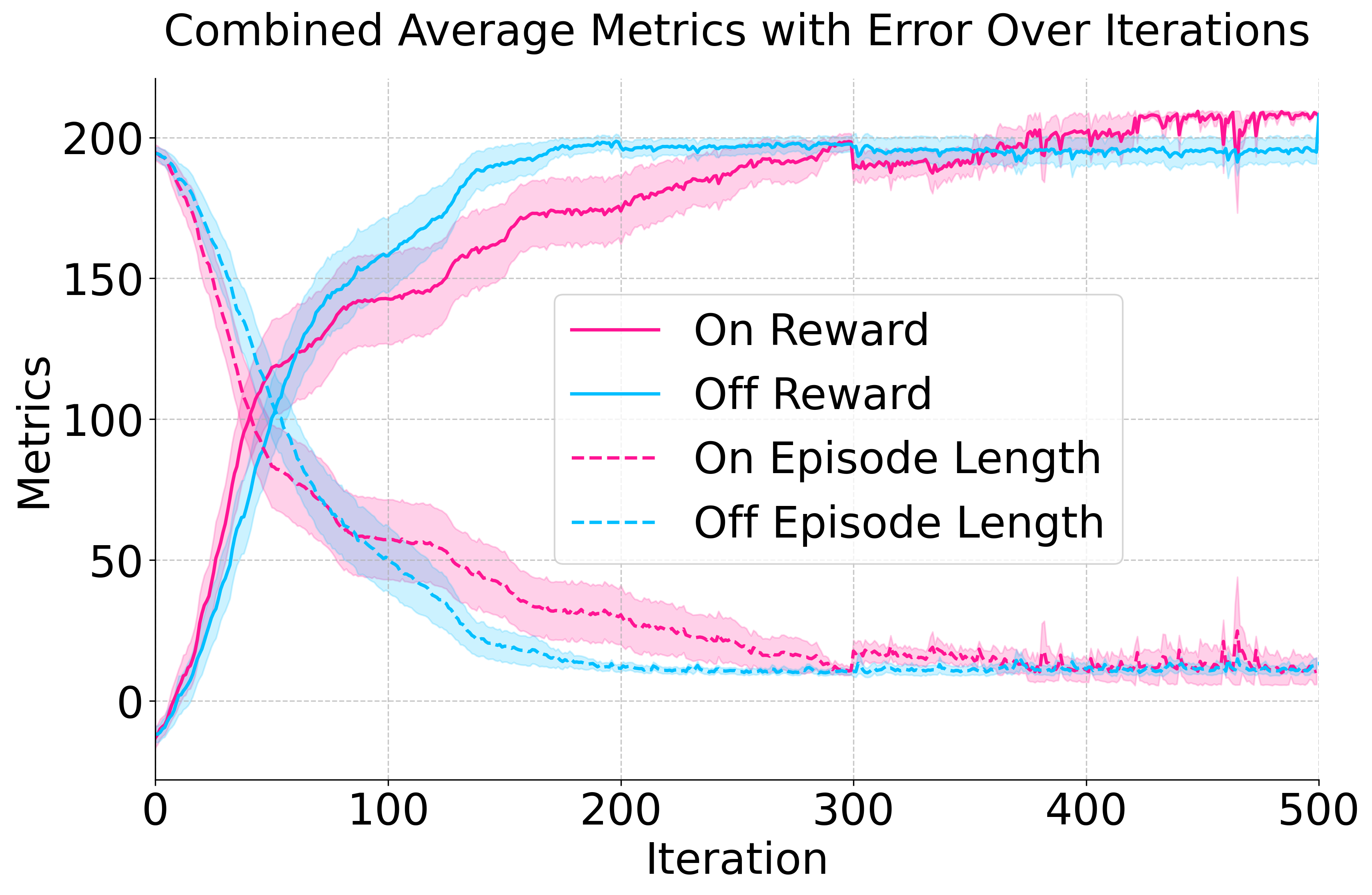}
	\caption{Average rewards and episode lengths over training iterations .}
	\label{fig:fig4}
\end{figure}

For each design scenario in Table \ref{tab:scenario_data}, we trained a PPO agent using dynamic planning with both on-light and off-light transformations, utilizing identity-full walls in both cases. We then evaluate the layouts generated by our RL agents based on their adherence to the desired geometric properties (areas and aspect ratios) and topological requirements (adjacencies). Additionally, we consider factors such as convergence speed. Figure \ref{fig:fig4} illustrates the learning curves for the agent using on-light and off-light transformations in SLD. Both approaches show consistent improvement over time, with rewards increasing and episode lengths decreasing. 

\begin{figure}
	\centering
	\includegraphics[width=\linewidth]{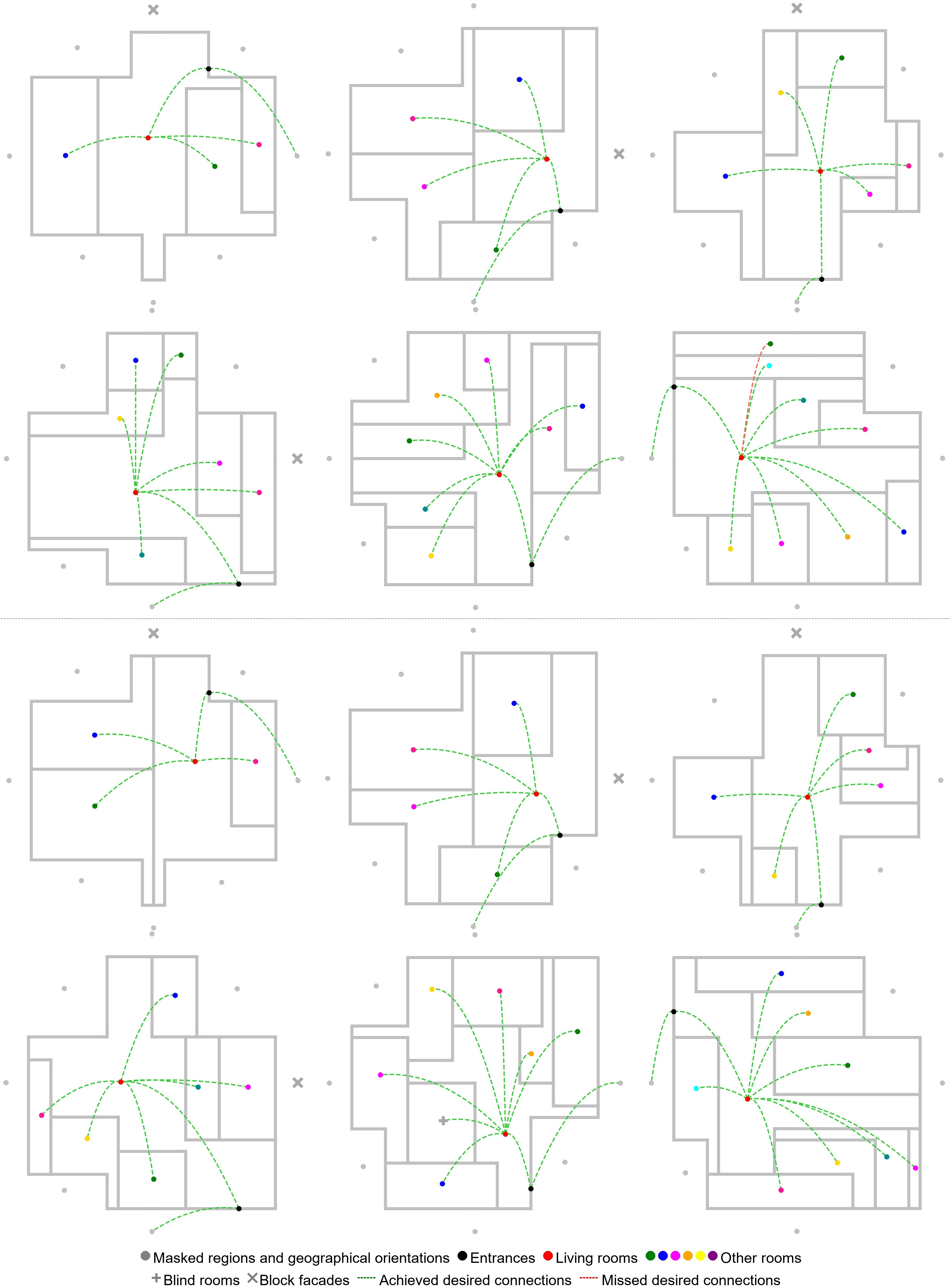}
	\caption{Comparison of space layouts generated using off-light (top two rows) and on-light (bottom two rows) transformations.}
	\label{fig:fig5}
\end{figure}

The layouts generated by the trained PPO agents are shown in Figure \ref{fig:fig5} provide a clear visual representation of various design elements. Geometrically, the agent achieves a high degree of precision in meeting room area and aspect ratio requirements. The average difference between desired and achieved room areas across all scenarios is minimal, and the achieved aspect ratios closely match the specified values (both less than $5\%$). Topologically, the agent excels in establishing the required connections between rooms and living rooms and facades. The green dashed lines in each layout represent successfully implemented adjacencies, while the red dashed line and gray plus sign indicate instances where the agent failed to create a requested connection. Across all scenarios, there are $72$ potential connections, calculated as the sum of room-to-living room and room-to-facade connections. The agent successfully establishes all but $2$ of these connections, showing its proficiency in meeting adjacency requirements.

The results underscore the effectiveness of our RL-based approach in generating space layouts that align closely with specified design criteria. The agent demonstrates a strong capacity for balancing multiple objectives simultaneously, including room sizing, aspect ratios, and inter-room connectivity, while adapting to different levels of complexity in the design scenarios. This performance suggests that our method could be a valuable tool for architects and designers, offering efficient and creative solutions to complex space layout problems.\\\\

\subsubsection{Design trajectory}
\begin{figure}
	\centering
	\includegraphics[width=\linewidth]{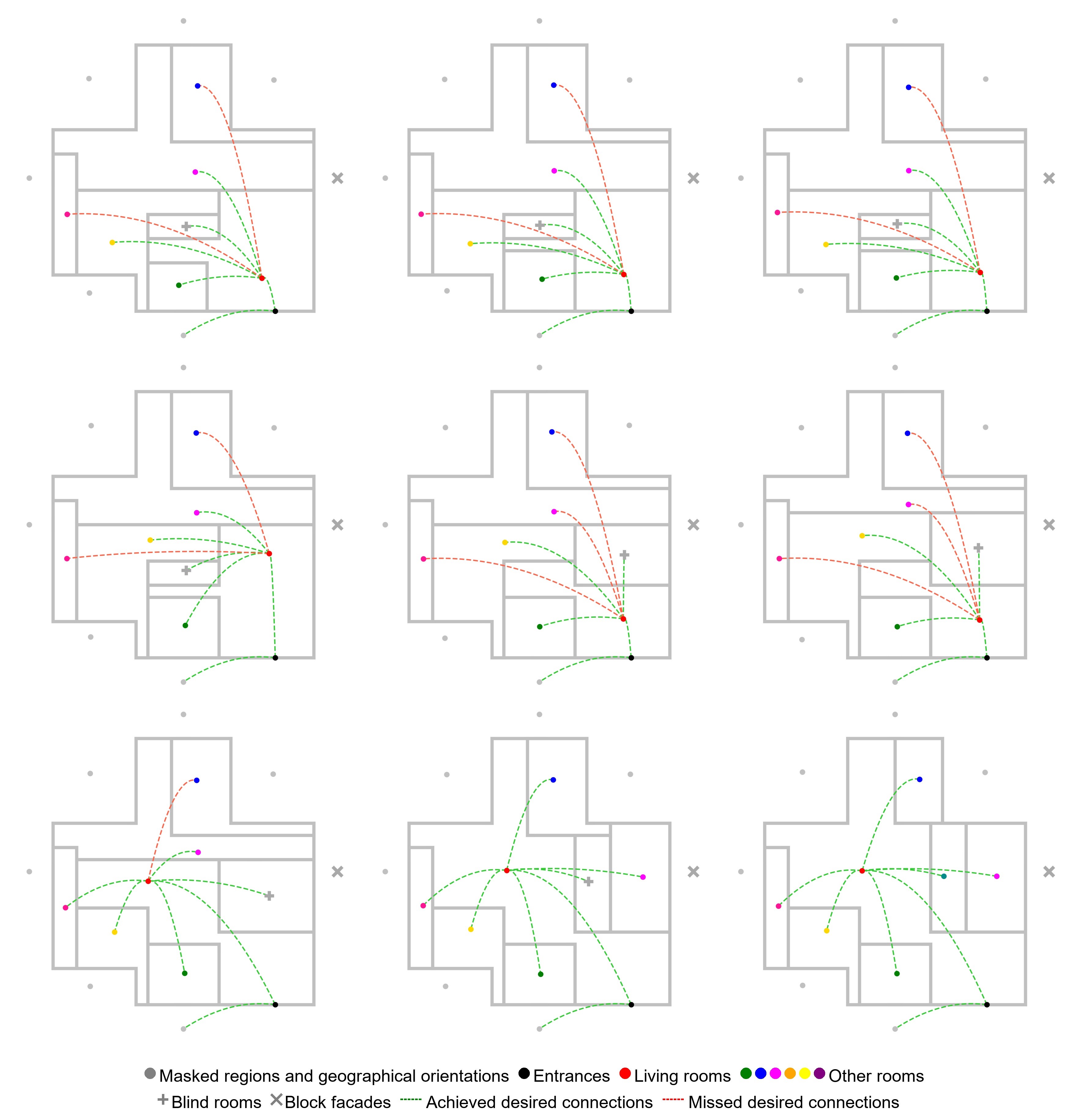}
	\caption{Agent's action trajectory from initial random layout to final optimal layout.}
	\label{fig:fig6}
\end{figure}

To understand the trained agent's behavior, we illustrate the action trajectory in Figure \ref{fig:fig6} which shows the step-by-step transformation of an SLD by our trained RL agent for scenario $4$. Starting from an initial random configuration, each subsequent image represents the layout after an action taken by the agent. Key observations include the gradual improvement in room proportions, the establishment of required adjacencies, and the overall organization of space becoming more coherent. This visual trajectory not only showcases the effectiveness of our RL approach but also provides insights into the decision-making process of the agent in the dynamic planning, highlighting how it balances various design criteria to achieve the final, optimized layout that satisfies the geometrical and topological properties defined in the design scenario.

\section{Discussion and Conclusion}
Our research demonstrates the effectiveness of the proposed laser-wall partitioning method in SLD. We framed the layout design problem as an MDP and successfully trained PPO agents to explore the solution space to find the optimal solution. The experiments across various scenarios highlight the adaptability and efficiency of our partitioning method in generating layouts that satisfy complex geometric and topological requirements. The comparison between straight and angled walls reveals the importance of flexibility in composing the space. Successful training of RL agents in different on-light, off-light dynamic planning emphasizes the robustness of the  proposed laser-wall method. The visual trajectory of the agent's decision-making process provides insights into how it navigates the solution space, progressively refining layouts to meet specified criteria.

While our approach shows promise, there are areas for future exploration. Integrating more architectural constraints, such as wall alignments and room width and length constraints, could further enhance the practical applicability of this method. Additionally, exploring the potential of transfer learning between different design scenarios could improve the efficiency and generalization capabilities of the RL agent. Comparing dynamic planning and one-shot planning is also left for future research. As we continue to refine this approach, the ultimate goal is to develop a tool that not only assists architects in generating initial design concepts but also serves as a collaborative partner in the creative process, offering novel solutions that human designers might not have considered. We already open sourced the SpaceLayoutGym simulator to support open research in architectural design and RL:  
\href{https://github.com/RezaKakooee/space_layout_gym}{SpaceLayoutGym}

\bibliographystyle{unsrtnat}
\bibliography{references}  

\end{document}